\documentclass{article}

\usepackage{arxiv}

\usepackage[utf8]{inputenc} 
\usepackage[T1]{fontenc}    
\usepackage{hyperref}       
\usepackage{url}            
\usepackage{booktabs}       
\usepackage{amsfonts}       
\usepackage{nicefrac}       
\usepackage{microtype}      
\usepackage{lipsum}
\usepackage{graphicx}
\usepackage[numbers]{natbib}
\usepackage{hyperref}
\usepackage{subcaption}
\usepackage{appendix}
\usepackage{xcolor}
\usepackage{mathtools}
\graphicspath{ {./images/} }

\title{Traffic forecasting on traffic movie snippets \vspace{0.5em}\\ \large 
Traffic4cast 2021 extended challenge}

\author{
 Nina Wiedemann \\
  Institute of Cartography and Geoinformation\\
  ETH Zurich\\
  \texttt{nwiedemann@ethz.ch} \\
 \And
  Martin Raubal \\
  Institute of Cartography and Geoinformation\\
  ETH Zurich\\
  \texttt{mraubal@ethz.ch} \\
}

\begin{document}
\maketitle
\begin{abstract}
Advances in traffic forecasting technology can greatly impact urban mobility. 
In the \textit{traffic4cast} competition, the task of short-term traffic prediction is tackled in unprecedented detail, with traffic volume and speed information available at 5 minute intervals and high spatial resolution.
To improve generalization to unknown cities, as required in the \textit{2021 extended challenge}, we propose to predict small quadratic city sections, rather than processing a full-city-raster at once. 
At test time, breaking down the test data into spatially-cropped overlapping snippets improves stability and robustness of the final predictions, since multiple patches covering one cell can be processed independently. 
With the performance on the \textit{traffic4cast} test data and further experiments on a validation set it is shown that patch-wise prediction indeed improves accuracy.
Further advantages can be gained with a Unet++ architecture and with an increasing number of patches per sample processed at test time.
We conclude that our snippet-based method, combined with other successful network architectures proposed in the competition, can leverage performance, in particular on unseen cities.
All source code is available at \url{https://github.com/NinaWie/NeurIPS2021-traffic4cast}.
\end{abstract}

\section{Introduction}

The transportation sector accounts for around 20\% of CO\textsubscript{2} emissions worldwide; out of which 41\% are due to vehicles. Traffic congestion in cities, with stop-and-go driving behavior, contributes substantially to these emissions~\cite{bharadwaj2017impact, barth2008real}. Nowadays, navigation systems or apps such as Google Maps have a great impact on routing decisions, and could therefore help to avoid congestion formation. However, such systems rely on projections into the future to assess the traffic state at the time of driving. In the IARAI \textit{traffic4cast} competition, the accurate prediction of traffic volume and speed is tackled in a novel approach, transforming sensor data to video-like 3D rasters in space and time. Since the first edition of the challenge in 2019, various methods have been proposed and were summarized and analysed by the organisers~\cite{kreil2020surprising, kopp2021traffic4cast}. It was found that the U-Net~\cite{ronneberger2015u}, an encoder-decoder style neural network architecture, that was initially developed for medical image segmentation, achieves better performance than other tested architectures~\cite{choi2020utilizing, choi2019traffic, martin2019traffic4cast}, despite conflating the temporal dimension of the one-hour input data into image channels. Furthermore, it was shown that Graph Convolutional Networks (GCNs) can improve the generalization to unknown cities~\cite{martin2020graph}. The 2021 \textit{traffic4cast} challenge further aims to test such generalization abilities in a newly introduced \textit{extended challenge}. In this challenge, a model is required to predict traffic in two unseen cities, based on training data from other cities.

Importantly, all cities are standardised into a $495\times436$ raster grid, and the baseline methods for \textit{core-} and \textit{extended} challenge process the data in this format. However, the underlying street maps differ, raising the question of whether feeding such high dimensional city-wide data to the model is actually necessary, or even hindering the transfer to other street layouts. In this work, we aimed to explore the effect of sub-dividing the problem by forecasting traffic on smaller sections of the city, or in other words "patches" of the original grid. We hypothesize that dividing the input data and merging the predictions will improve in particular the performance in the \textit{extended challenge}. Specifically, we train segmentation models on randomly cut $d\times d$ snippets of the training data, where the temporal and channel dimensions are left intact. At test time, we convolve the test data and predict the output for these regularly-spaced overlapping patches (see \autoref{fig:overview}). Crucially, the method yields an ensemble in itself since it outputs multiple predictions for each cell. In our experiments we find that 1) sub-division indeed improves the performance on unseen data, 2) accuracy increases slightly with smaller convolution "stride" at test time, and 3) using a U-Net++ on the patches is superior to a U-Net. In the following, the methods and results will be presented in more detail, and discussed with respect to related works.



\section{Methods}

\begin{figure}
    \centering
    \includegraphics[width=\textwidth]{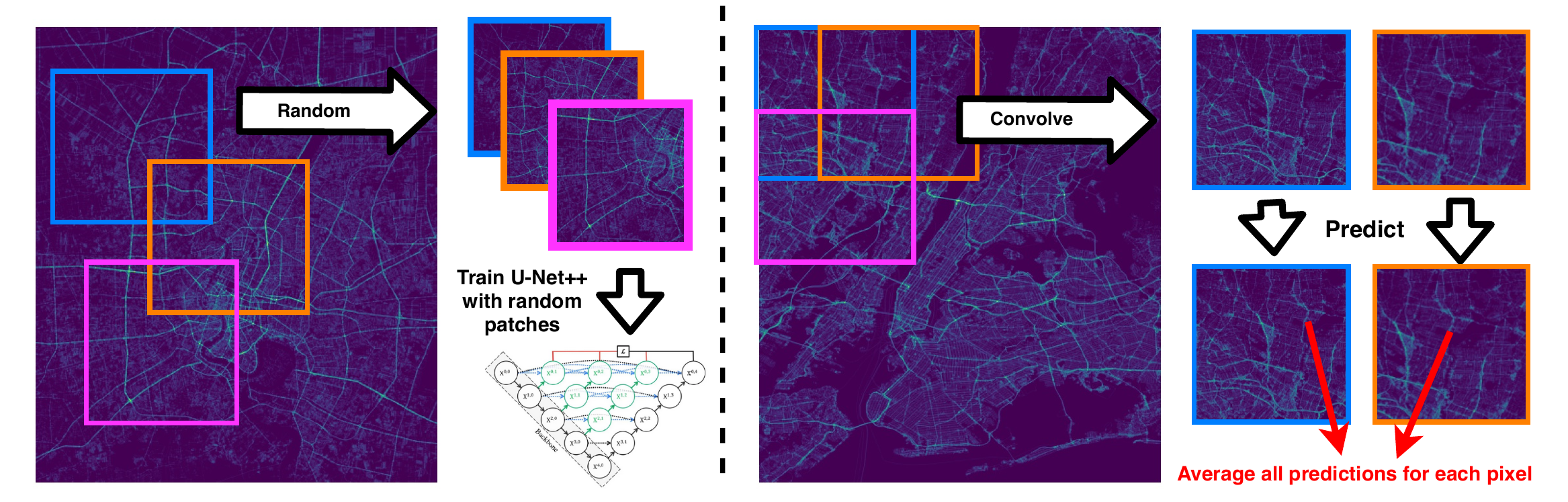}
    \caption{At \textbf{train time (left)}, random quadratic samples are drawn from the training cities. \textbf{At test time (right)}, overlapping patches are extracted from the unseen city and they are predicted independently. The predictions per pixel are averaged.}
    \label{fig:overview}
\end{figure}

Our source code, which led to the third rank on the leaderboard, only required a small set of changes to the code base provided by the organizers. Mainly, we added Unet++ to the available models, and integrate the necessary dataset structure to enable patch-based processing. An overview of the method is shown in \autoref{fig:overview}.

\subsection{Data preprocessing}
The dataset of the 2021 competition consists of 10 cities that are compressed into $495\times 436$ rasters. The data constitutes traffic volume and average speed values at 5-minute intervals, normalised to values between 0 and 255. Volume and speed values are encoded in four channels each, corresponding to the four headings NE, SE, SW, NW. The data is given in HDF5 files; one file covers one day (288 time slots) and is therefore a four-dimensional raster of size $288 \times 495 \times 436 \times 8$. The task is to predict speed and volume up to one hour into the future, given the previous hour as input. For further details on the data see a summary of the 2020 competition~\cite{kopp2021traffic4cast} or the conference website~\footnote{\url{https://www.iarai.ac.at/traffic4cast/}}.

We implement a new Pytorch dataset class where data are cached in regular steps. At each step, $m$ training files are loaded, then $k$ snippets per file are sampled, and these $m\cdot k$ samples are normalized and reshaped. In the baseline source code, the files are dynamically loaded during training. However, this becomes inefficient for smaller samples, when data loading takes longer than the backpropagation step. Since each file covers a full day, samples that are diverse in space and time can be drawn from a single file. Here, we sample $k=10$ patches and $m=100$ files in each step. The cached dataset is renewed every two epochs, such that all "cached" patches are seen twice by the model before a new dataset is loaded. After each epoch, we evaluate the results on an independent validation set of $10$ files ($10$ patches each). To gain a good estimate on the generalization ability of the model, we use the data from an entire city as the validation set, and exclude this city from the training data for our experiments. Furthermore, the inputs are normalized to values between 0 and 1 as it is commonly done in the computer vision literature. Last, the input raster is padded with 6 rows/columns of zeros at all sides, and the temporal dimension is transformed into the channel representation, leading to 96 values per cell for 12 time slots with 8 channels. In summary, our preprocessing steps yield input data of size $96 \times (6+d+6) \times (6+d+6)$, where $d$ denotes the side length of the quadratic patches.

\subsection{Training}
We use a U-Net++~\cite{zhou2018unet++, zhou2019unetplusplus} to forecast the traffic volume and speed for the subsequent hour. U-Net++ is a successor architecture to the U-Net that won previous editions of the \textit{traffic4cast} challenge, and it was shown superior to the original U-Net in related tasks~\cite{alexakis2020evaluation}. In the \textit{traffic4cast 2020} challenge, \citet{qi2020traffic4cast} already noticed better performance with Unet++, but they did not adopt it in their solution due to its larger number of parameters. Unet++ is characterised by more dense skip connections, as well as further convolutional layers on the skip pathways. Due to the sparsity of the data, we believe that additional skip connections could improve performance. In addition, we propose to use a Sigmoid activation in the final layer in order to output values between 0 and 1. This accelerates training since the model is not required to learn an output range between 0 and 255.

We train our model for 1000 epochs, where one epoch consists of 1000 ($m\cdot k = 100\cdot 10$) patches as described above. A batch size of 8 can be used due to the lower resolution of the data. As in the U-Net baseline provided by the organizers, a mean squared error loss function is used and the model is trained with Adam Optimizer. All models are trained on a single TITAN RTX GPU. Finally, it was found that prediction accuracy improved slightly with checkpoint averaging; i.e. the weights of the saved checkpoints of the last 200 time steps are averaged.

\subsection{Merging patch-wise predictions}

At test time, the data of a new city is split into patches by convolving the raster in regular steps of size $s\ (s<d)$. For example, if $d=100$ and $s=50$, each cell is contained in four patches, except for cells at the border (see \autoref{fig:overview} right). Since the original size ($495\times 436$) can not be divided into regularly-spaced patches of the same size, a patch for the last column $(x,436-d)$ and last row $(495-d, y)$ is added. The predictions for all patches are merged by averaging the per-cell predictions. Note that the parameter $s$ is only relevant at test time since the training data is sampled randomly rather than in a regular grid. 

\section{Results}
%

\subsection{Evaluation of model and parameter choices}
The increase in performance due to splitting the data into patches, as well as replacing the U-Net by a Unet++, led our approach to rank third in the extended challenge of \textit{traffic4cast 2021}. \autoref{tab:results} lists the results of submitted predictions for the test data, with the scores as displayed on the challenge leaderboard. While the absolute differences of mean squared error (MSE) might seem small, the ranks on the leaderboard were shown to be significant by~\cite{kreil2020surprising}, at least for the 2019 challenge. Our best result was obtained with a Unet++ and patch-wise training with $d=100$. With the same model we also obtained our best score in the \textit{core challenge}, namely a MSE of 50.25. This result indicates that splitting the raster into smaller parts is beneficial not only for spatio-temporal generalization, but also for simple temporal prediction. Furthermore, in our experiments we consistently observed worse results with $d=60$, indicating that larger patches are necessary to provide sufficient spatial context for predictions up to one hour into the future.
\begin{table}
    \centering
    \resizebox{0.5\textwidth}{!}{
    \begin{tabular}[b]{cccc}
    \textbf{Model} & \textbf{Patch size (d x d)} & \textbf{Stride} & \textbf{Score (MSE)} \\
    \toprule
    U-Net          & 495 x 436               & -               & 64.2           \\
    Unet++         & 495 x 436               & -               & 60.93          \\
    U-Net          & 100  x 100                & 50              & 61.32          \\
    Unet++         & 100  x 100                & 50              & 60.134         \\
    Unet++         & 60  x 60                & 30              & 60.44          \\
    Unet++         & 100  x 100                & 10              & 59.93         
    \end{tabular}}
    \vspace{1em}
    \caption{Scores achieved on the competition leaderboard. Our best performing approach is a Unet++ operating on 100 x 100 video snippets.}\label{tab:results}
\end{table}
\begin{figure}[ht]
    \centering
    \begin{subfigure}[b]{0.55\textwidth}
    \includegraphics[width=\textwidth]{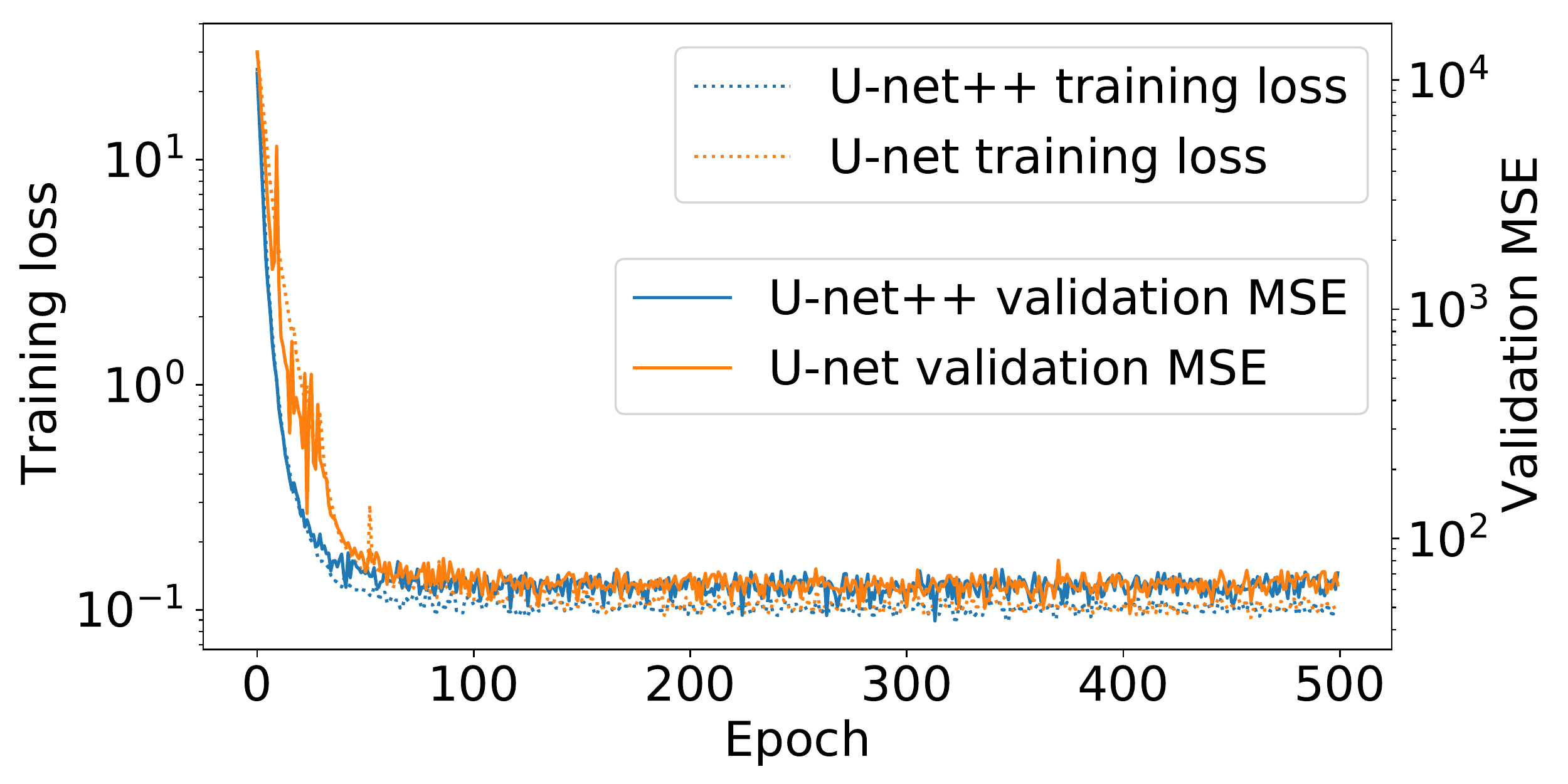}
    \caption{Loss convergence of U-Net compared to Unet++}
    \label{fig:unet}
    \end{subfigure}
    \hfill
    \begin{subfigure}[b]{0.42\textwidth}
    \includegraphics[width=\textwidth]{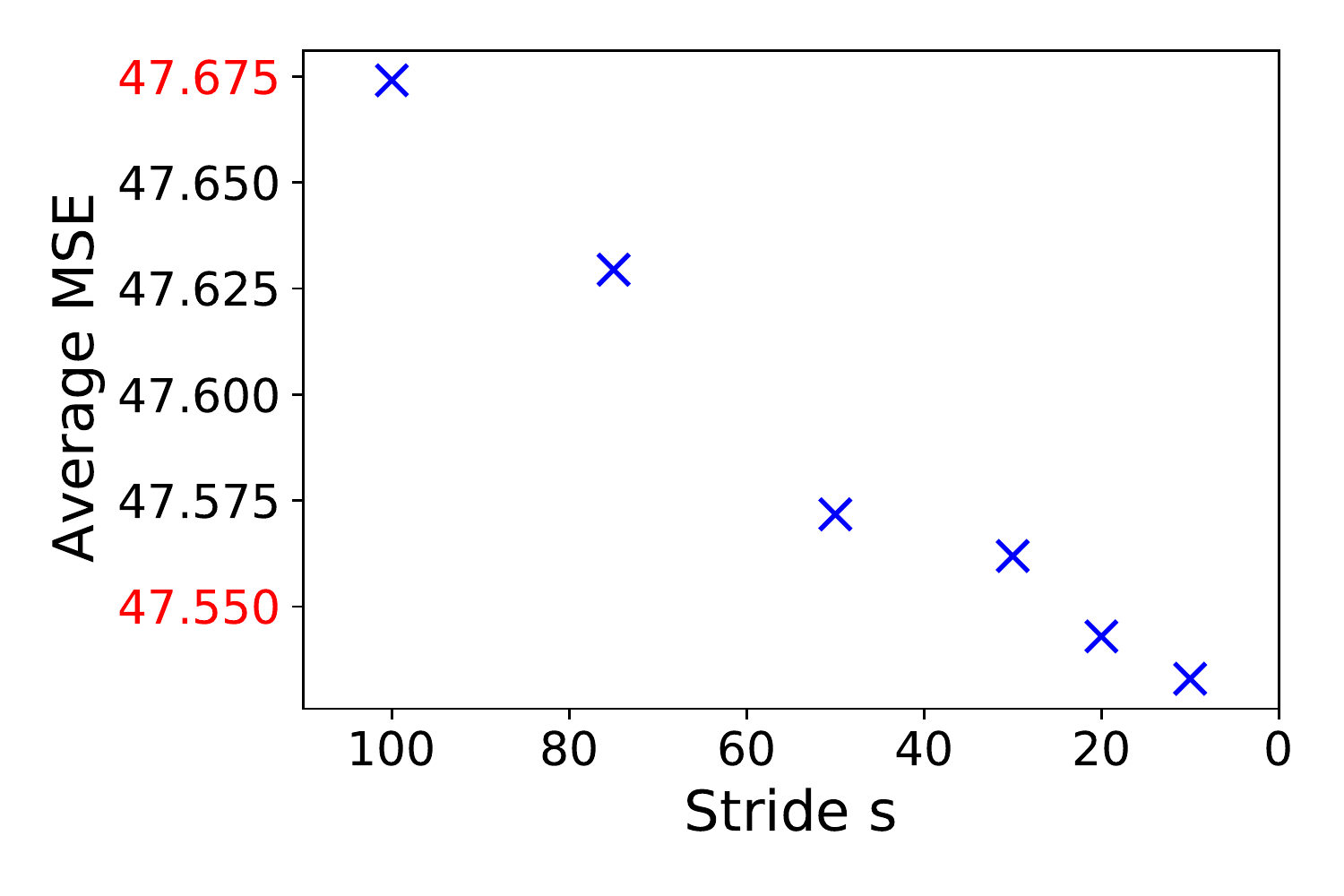}
    \caption{Advantages of lower stride}
    \label{fig:stride}
    \end{subfigure}
    \caption{Effects of model choice and stride parameter setting. 100 samples from the city Antwerp were used as a validation set. Unet++ converges faster and to marginally lower validation MSE (\subref{fig:unet}). At test time it is advantageous to convolve the raster in small steps (''stride''), but the effect is small, as seen in the red values (\subref{fig:stride}).}
    \label{fig:results}
\end{figure}

Further analysis was conducted using a validation set left out during training. After every epoch, 10 patches are sampled from 10 randomly selected files of the city Antwerp, and the MSE is computed. \autoref{fig:unet} shows the convergence of the training loss, as well as the validation MSE of the U-Net and Unet++. For this analysis we fix $d=100$, and evaluate the validation MSE on a collection of patches instead of merged predictions of one sample. Unet++ converges faster than the U-Net and also achieves a better MSE after 500 epochs. Taking the average over the last 20 epochs (\autoref{fig:unet}) yields an average MSE of 65.66 and 64.03 for U-Net and Unet++ respectively, confirming the differences in scores shown in \autoref{tab:results}.

Finally, we explore how much can be gained with decreasing the parameter $s$. The step size $s$ at test time can be thought of as the stride of a convolution, and it determines the number of predictions per cell at test time. If $d=100$, $s$ can at most be $100$ to ensure that the whole raster is covered. To select data that is representative of the test data, we use the additional metadata (weekdays and time) from Berlin, and sample data from Antwerp accordingly. For example, if a test sample for Berlin would be at time slot 130 on a Monday, we would select a random Monday-date from the available Antwerp data and extract the same time slot. Using our best model (Unet++, $d=100$), we predict the validation data with varying stride, namely $s\in \{10, 20, 30, 50, 75, 100\}$. \autoref{fig:stride} shows that the average MSE decreases with lower stride. However, only an improvement of $0.125$ is gained on average with $s=10$. Note that $s=10$ already leads to 1435 patches\footnote{When regular patches of size $d$ are extracted with stride $s$ from the original raster ($495\times 436$), the number of final patches is computed as $\Big(\lceil \frac{495 - d + s}{s}\rceil \Big) \cdot \Big( \lceil \frac{436 - d + s}{s}\rceil \Big)$ which is $41 * 35 = 1435$ for $d=100,\ s=10$.} per sample, and up to 121 predictions per cell, at the cost of a longer runtime. The result demonstrates that the ensemble-like behavior of patch-wise prediction accounts for a small but significant increase in performance, whereas the main advantage of the method however lies in the simplification of the problem by splitting the data into smaller parts.

\subsection{Error analysis and further approaches}

We further analyse the characteristics of the predictions from our best-performing model. Note that the following observations are yielded from a small validation set (10 files) and might not be representative for the performance on the test data. As the challenge organizers noted in their data exploration files, much of the error is due to predictions in the speed channels, with a MSE of 0.065 for the volume channels, and a MSE of 112 for the speed channels. The overall low MSE is however mostly due to the ability of the model to correctly predict zero activity: More than 98\% of the volumes in the validation set were zero, and out of those, 99.6\% were predicted correctly as zero. Considering this strong sparsity, we divide the speed-MSE further into the error on cells where the volume is null ($\text{MSE}_{\text{speed}}(\overline{V})$) and the speed-MSE where the volume is not null ($\text{MSE}_{\text{speed}}(V)$) in the ground truth. Due to the sparsity, the number of ground-truth null-volume-cells $|\overline{V}|$ is much higher than the number of cells with a positive volume $|V|$. 
%
We find that $\text{MSE}_{\text{speed}}(\overline{V}) \approx 35$ whereas $\text{MSE}_{\text{speed}}(V) \approx 10000$. Crucially, the latter is hardly better than chance. The overall MSE actually improves when all the speed predictions for all cells in $V$ are simply set to $127\ (= 0.5 \cdot 255)$. However, such change is not possible during test time, because the volume is itself predicted, so it is uncertain whether a cell is part of $V$ or $\overline{V}$. Accordingly, we observed that the predicted speeds are oftentimes smaller than the ground truth speeds, which can be explained by the uncertainty of the model with respect to the volume: If the model is uncertain whether the cell is in $V$ or in $\overline{V}$, a speed value of 30 will yield a better MSE in expectation than a speed of 127, at least for sparse data.

Based on these observations, we attempted to modify the corresponding speed values, or to set volume and speed to zero based on the underlying street network, but did not gain any improvement. We further tried to modify the loss function accordingly, similar to attempts of \citet{liu2019building} in the 2019 competition. For example, the model was trained to only predict the speed \textit{on non-zero ground truth volumes}, masking the other values in the loss function. At test time, we  then set all speeds to zero where the predicted volume was zero. Although the results were not satisfactory (MSE>70) we believe such modified loss functions could be more appropriate for the data and might be an interesting endeavour for future research. Further approaches that were explored during the competition are explained in the appendix.

\section{Discussion}
Our third-place winning approach of the \textit{traffic4cast 2021} extended challenge led to the following insights: We have provided evidence that subdividing city-wide rasters into smaller patches improves the prediction performance, both on seen and on unseen cities. The prediction of overlapping patches at test time yields an ensemble-type output, where the error decreases with an increasing variety of patches covering each cell. Meanwhile, the computational effort is lower due to smaller input data size, and the runtime for training is reduced. We believe that best performance could possibly be obtained when the winning model of the challenge, or an ensemble such as the solution proposed in 2020~\cite{choi2020utilizing}, is trained on patch-data as proposed here. The presented results also hint at the possibility that cropping all cities into a grid of the same size might not be advantageous in general and could be omitted in favour of maintaining the real city extents. 

Based on our error analysis, we see opportunities for further work in the development of tailored loss functions or post-processing methods. The prediction of speed should be disentangled from the prediction of volume to avoid the bias to predict low speed. Also, the underlying city map should be utilised to adjust the predictions based on available metadata. Other research directions could further explore how to better deal with the temporal dimension of the data. While LSTMs have not outperformed U-Net architectures in previous attempts, Temporal Convolutional Networks~\cite{lea2017temporal} or temporal graph networks~\cite{yu2017spatio, rossi2020temporal} might be a promising research direction.\\
Improvements in the predictive power of traffic analysis systems could greatly impact decisions in our daily life, and, on a macro-level, also determine the contribution of the transport sector to climate change. The \textit{traffic4cast} competition offers an excellent opportunity to explore deep learning methods for such technologies, and it complements other efforts (see~\cite{kumar2021applications}) to continuously incorporate ongoing research in deep learning and computer vision into the mobility field.

\bibliographystyle{plainnat}
\bibliography{references}

\begin{appendix}
\section*{Appendix}
\section{Testing other network architectures}

Apart from the main approach described above, various other methods were tried and should be noted here for further work on the \textit{traffic4cast} dataset. First, we explored other architectures with the Python package \texttt{segmentation\_models\_pytorch}~\cite{Yakubovskiy:2019}. Specifically, we trained their implementation of the Unet++, where the encoder is a ResNet model pretrained on ImageNet. The results were similar but slightly inferior to our Unet++ implementation. Furthermore, the Pyramid Scene Parsing Network~\cite{zhao2017pyramid}, the Pyramid Attention Network~\cite{li2018pyramid} and the DeepLabV3~\cite{chen2017rethinking} model were tested, but they converged to an MSE of more than 100. We believe that these models, at least with the current parameters, could not cope with the sparsity of the data and predicted values close to zero everywhere.

\section{Naive approach to temporal prediction}
Finally, we also implemented a naive baseline for the temporal challenge based on historic traffic volume. Since traffic is most of the time highly regular and depends mostly on the weekday and time, we hypothesised that optimal results could be achieved when setting the output to its expected values. Moreover, a method was developed to shift from the pre-COVID training data of 2019 to the COVID-affected traffic in 2020, explained in the following. We take advantage of the fact that we not only know the time of the day and weekday of the test data, but also the traffic of the previous hour. Let $0\leq t\leq 288$ be the time of the day in 2020 when the test period of one hour starts. Furthermore, let $A(t, w, y, c)$ denote the average traffic volume and speed at time $t$, weekday $w$, year $y$ and city $c$. Now we aim to predict the traffic for a test sample of city $\hat{c}$ at time $\hat{t}$, $\hat{w}$. The traffic of the previous hour, which we define as $T(\hat{t}-12, \hat{w}, 2020, \hat{c})$ is known. We can thereby quantify the day- and time-specific change from 2019 to 2020 with the ratio $$\frac{T(\hat{t}-12, \hat{w}, 2020, \hat{c})}{A(\hat{t}-12, \hat{w}, 2019, \hat{c})}$$
Last, the final result is computed by applying this shift to the average data of the test time slot:
$$
T(\hat{t}, \hat{w}, 2020, \hat{c}) = A(\hat{t}, \hat{w}, 2019, \hat{c}) \cdot
\frac{T(\hat{t}-12, \hat{w}, 2020, \hat{c})}{A(\hat{t}-12, \hat{w}, 2019, \hat{c})}
$$

\end{appendix}

\end{document}